\documentclass[preprint,journal]{vgtc}                     

\onlineid{}

\preprinttext{This manuscript is a preprint, and is under consideration for submission to IEEE transactions in future. }

\vgtccategory{Research}

\vgtcpapertype{technique}

\title{Instance-guided Cartoon Editing with a Large-scale Dataset}

\author{%
  Jian Lin,
  \authororcid{Chengze Li}{0000-0002-1519-750X},
  \authororcid{Xueting Liu}{0000-0002-0868-5353} and
  Zhongping Ge
}

\authorfooter{
The authors are with Caritas Institute of Higher Education, Hong Kong SAR, China.
E-mail: {jlin, czli, tliu, zge}@cihe.edu.hk
}

\abstract{%
 Cartoon editing, appreciated by both professional illustrators and hobbyists, allows extensive creative freedom and the development of original narratives within the cartoon domain. However, the existing literature on cartoon editing is complex and leans heavily on manual operations, owing to the challenge of automatic identification of individual character instances. Therefore, an automated segmentation of these elements becomes imperative to facilitate a variety of cartoon editing applications such as visual style editing, motion decomposition and transfer, and the computation of stereoscopic depths for an enriched visual experience. Unfortunately, most current segmentation methods are designed for natural photographs, failing to recognize from the intricate aesthetics of cartoon subjects, thus lowering segmentation quality. The major challenge stems from two key shortcomings: the rarity of high-quality cartoon dedicated datasets and the absence of competent models for high-resolution instance extraction on cartoons. To address this, we introduce a high-quality dataset of over 100k paired high-resolution cartoon images and their instance labeling masks. We also present an instance-aware image segmentation model that can generate accurate, high-resolution segmentation masks for characters in cartoon images. We present that the proposed approach enables a range of segmentation-dependent cartoon editing applications like 3D Ken Burns parallax effects, text-guided cartoon style editing, and puppet animation from illustrations and manga.
}


\teaser{
  \centering
    \includegraphics[width=\linewidth]{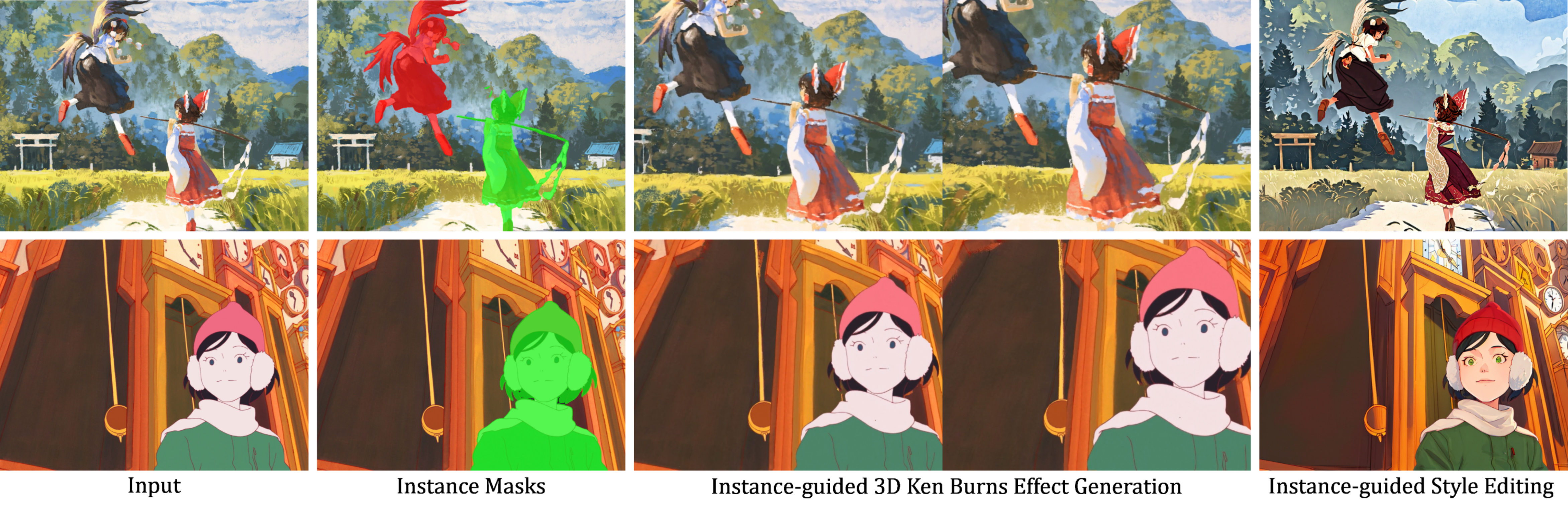}
    \caption{Given an input cartoon image, our method extracts high-quality instance masks for the subjects in the image. The extracted instance masks enable a variety of instance-guided cartoon applications, such as 3D Ken Burns effect and cartoon style editing.}
    \label{fig:teaser}
}

\renewcommand{\manuscriptnotetxt}{}

\nocopyrightspace

\usepackage{tabu}                      
\usepackage{booktabs}                  
\usepackage{lipsum}                    
\usepackage{mwe}                       
\usepackage{amsmath}
\usepackage{makecell}
\usepackage{mathabx}

\usepackage{mathptmx}                  

\begin{document}


\firstsection{Introduction}

\maketitle

\label{sec:intro}
Cartoon has been a popular art form enjoyed by people all over the world. Along with its popularity, cartoon editing has garnered significant interest among both professional illustrators and hobbyists, enabling unprecedented freedom in content creation and expansion of original narratives within the cartoon domain. For example, for enriched visual experiences, it is possible to apply 3D Ken Burns effect or movement effect to create a cartoon clip from a single cartoon image. It is also possible to translate the styles of the input cartoon image to other cartoon styles. An example is shown in Fig.~\ref{fig:teaser}. To apply these editing operations, the illustrators need to first segment out the subjects carefully from the image and fill in the missing pixels of the background, and then apply the above effects based on segmentation and inpainting. Unfortunately, the manual segmentation process is not only tedious and time-consuming, but also error-prone when the boundary of the object contains delicate structures or is semi-transparent. An automatic segmentation method to acquire precise instance masks is highly desired.

Current learning-based segmentation solutions, despite their numerous advancements~\cite{maskrcnn,lyu2022rtmdet,groundedsegmentanything}, are primarily attuned to visual recognition within natural images. These approaches typically operate on a relatively lower resolution scale, which suffices for recognition tasks but falls short in supporting high-quality post-processing and editing. In stark contrast, the processing of cartoon images requires a high level of segmentation quality. Cartoons, characterized by explicit structural line inking, demand precision that extends beyond the capabilities of conventional techniques. Any compromise on the quality of segmentation can lead to severe visual artifacts during the post-segmentation editing stage, such as incorrect estimations in depth and poses, aliasing artifacts, etc. Consequently, applying these traditional methodologies to cartoons often leads to undesirable results. As depicted in Fig.~\ref{fig:qualitative_evaluation}, these techniques typically yield incomplete extraction of subject instances or induce artifacts in the segmentation masks of each extracted instance. This outcome clearly falls short of our objective for high-quality post-segmentation processing and editing of cartoons. This gap can be attributed to the relative absence of cartoon images in the training datasets and the lack of appropriate models specifically designed for detailed high-resolution instance extraction.

To acquire high-quality segmentation results for cartoon images of complex anime scenes involving multiple characters, we first propose a large-scale cartoon segmentation dataset consisting of 98.6k paired high-resolution cartoon images with their ground truth subject instance labeling masks. This data set is prepared using a novel reverse engineering approach that extracts character and object instances from chroma-keying videos (Fig.~\ref{fig:chroma_key}) and still illustrations, and simulates the actual composition of cartoons and animations through synthesis. This large-scale cartoon segmentation dataset serves as the basis for accurate subject segmentation from cartoon images of complex anime scenes involving multiple characters. Furthermore, to enable high-quality subject segmentation from cartoon images, we propose a novel instance-aware image segmentation model that can generate precise, high-resolution segmentation masks. This model consists of two stages, a low-resolution segmentation stage to locate a rough but occlusion-free mask for each subject instance, and a high-resolution segmentation stage to obtain a high-resolution instance mask for each subject under the guidance of the rough instance mask acquired in the first stage. The extracted high-resolution masks enable a variety of segmentation-dependent cartoon editing applications, including but not limited to 3D Ken Burns parallax effects, text-guided cartoon style editing, and puppet animation from illustrations and manga.

We evaluate our methods on a large variety of cartoons, illustrations, and even manga. Our method achieves a significant superiority over the existing methods in both qualitative and quantitative evaluations. Three applications are presented to demonstrate the boosts of cartoon editing effects based on the extracted high-quality instance segments. Our contributions are summarized as follows.
\begin{itemize}
    \item To the best of our knowledge, this work is the first attempt at a high-resolution subject instance segmentation method for cartoons.
    \item We propose a large-scale cartoon segmentation dataset prepared via a novel reverse engineering approach based on foreground subjects extracted from chroma-keying videos and illustrations.
    \item Our extracted high-quality instance segmentation significantly streamlines various segmentation-dependent cartoon production pipelines.
\end{itemize}

\section{Related Work}
\label{sec:related}
\subsection{Image and Instance Segmentation}

Instance segmentation identifies and labels individual objects in an image, offering an advantage over semantic segmentation by distinguishing different instances of the same object class. This step is crucial for cartoon editing, where the recognition of individual subjects is substantial to various applications, such as scene understanding, style and motion editing, etc. 
Among various instance segmentation methodologies, Mask R-CNN~\cite{maskrcnn} extends Faster R-CNN~\cite{fasterrcnn} to predict individual object masks in parallel with the existing branch for bounding box recognition. Nevertheless, this method only produces low-resolution instance masks, and the overall computational complexity is relatively high. 
Recent advances such as model architecture optimization, improved training strategy, and mask head design enhancement contribute to better performance in instance segmentation networks. Among these methods, RTMDet-Ins~\cite{lyu2022rtmdet} is considered one of the state-of-the-art methods, which uses an efficient backbone, CSPDarkNet~\cite{bochkovskiy2020yolov4}, and an instance-aware mask head~\cite{CondInst2022Tian} that dynamically generates convolution kernels conditioned on target instances. This model does not require RoI proposals, which significantly improves performance. However, despite their merits, these instance segmentation networks generally predict masks at a low resolution, leading to coarse, polygon-like instance masks, which can hardly be used for cartoon editing tasks which requires precision in extracting instance boundaries.

The recent advancement of Segment Anything (SAM)~\cite{sam}, a promptable segmentation model, has significantly improved the quality of instance segmentation. Trained with more than a billion masks on 11 million licensed, privacy-respecting images, SAM can generalize across diverse media types, including cartoons and manga. Despite this, SAM requires additional detector inputs (e.g., an anchor point, a bounding box) for initialization, preventing fully automatic segmentation. Grounded SAM~\cite{liu2023grounding}~\cite{groundedsegmentanything} attempts to overcome this limitation by extending the SAM architecture with a two-stage design. They infer initial bounding boxes for SAM from text input, with the aim of achieving prompt-based automatic instance extraction. Nonetheless, its primary use on natural photographs may result in inaccurate segmentations for cartoons and animations. The quality of its segmentation results is tightly related to the initial extraction stage, potentially leading to missed instances or false positives. The segmentation model YAAS~\cite{yaas}, derived from CondInst~\cite{CondInst2022Tian} and SOLOv2~\cite{wang2020solov2}, was introduced with a specific dataset of 945 cartoon/anime images. Yet, due to the scarcity of high-quality cartoon data, it struggles with accurate instance mask prediction and understanding complex multi-character anime scenes.
On the contrary, our method adopts sophisticated model frameworks and a novel dataset of 98.6k cartoon and anime samples, which successfully generates high-quality segmentation for high-resolution input, thereby yielding superior results.

Matting techniques possess the potential for high-quality foreground object extraction but typically require an initial manual guess via a confidence-based trimap, posing challenges for automatic instance extraction. Trimap-free matting methods~\cite{ke2022modnet, li2022bridging} remove the trimap requirement, yet focus on foreground-background separation, not individual instances. Sun et al.~\cite{sun2022human} uses a pretrained Mask R-CNN for trimap generation and instance matting, further refined using aggregated features. This method is trained on a mixture of natural and synthetic data and thus has limited generalizability to cartoons. Furthermore, it depends on a pretrained Mask R-CNN which was proved to be ineffective on cartoon images.

\subsection{Datasets for Cartoon Segmentation}

Datasets are essential for deep learning methods, and particularly so in the field of cartoon editing due to the domain gap between natural photographs and cartoons. Existing models, such as those trained on the MSCOCO dataset~\cite{MSCOCO}, often perform poorly when applied to cartoons. To address this, a cartoon-focused dataset is needed. However, currently there is an absence of large-scale, high-resolution cartoon datasets with detailed instance labeling. AniSeg~\cite{aniseg} presents a small dataset of 945 character-mask pairs and a synthetic dataset generated from these pairs. The dataset is adopted by the YAAS model~\cite{yaas} mentioned before. However, the limited diversity of the data restricts the generalization ability of the models trained upon it, especially when the input images are featuring multiple character instances. 
Chen et al.~\cite{chen2022bizarre} offer a larger synthetic cartoon dataset, combining 18.5k foreground images from the Danbooru dataset~\cite{danbooru2021} with 8.1k background images from the Danbooru and Pixiv datasets. However, all foreground images in this dataset are still-frame illustrations, which have different visuals from actual cartoons and animations. Furthermore, this dataset is not publicly available.
In this work, we propose a new high-quality dataset with 98.6k cartoon image-mask pairs created from 65k foreground instances and 8k background scenic images. Especially, the foreground instances are sourced from our collected chroma-keying videos and the Danbooru dataset~\cite{danbooru2021} and the background images are sourced from both illustrations and real-world animations. High-quality segmentation masks can be achieved on the basis of this newly prepared dataset. Our dataset will be released upon acceptance of the article.

\begin{figure}[!t]
    \centering
    \includegraphics[width=\linewidth]{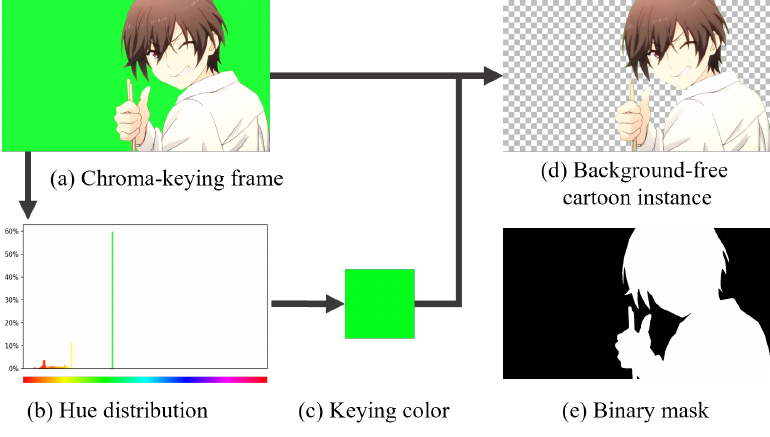}
    \caption{Keying-based foreground preparation. We estimate the keying colors based on hue distribution, to produce the background-free instance image and its corresponding mask.}
    \label{fig:chroma_key}
\end{figure}

\section{Cartoon Segmentation Dataset}
\label{sec:dataset}


\begin{figure}[!t]
    \centering
    \includegraphics[width=0.99\linewidth]{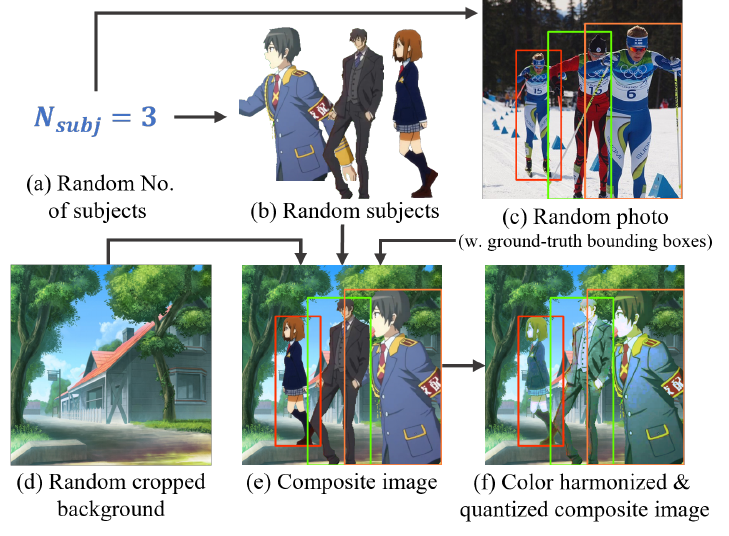}
    \caption{Procedures to synthesize our cartoon segmentation dataset. Specifically, the number of subjects are sampled from a Poisson distribution and each subject will be aligned to the closest bounding box guided by a labelled reference photo. Together with a random cropped background image, the composite image undergoes an optional post-processing step, to serve as a training image.}
    \label{fig:composition}
\end{figure}

The disparity between natural photos and cartoons hampers the performance of networks trained on natural photo datasets when applied to cartoons. For improved cartoon instance segmentation, a large-scale cartoon dataset with instance-level labels is necessary. However, due to the complexity of the task and the challenges in pixel-perfect labeling, manually creating such a dataset is not feasible. We propose a reverse engineering method to overcome this. This method extracts instances of characters and objects from chroma-keying videos and drawings and synthesizes them to replicate the actual cartoon and animation composition. Although reverse engineering is a common approach in dataset preparation, the synthesis of ground truths that mirror real cartoons is still demanding, especially considering the diversity of characters, poses, and compositions in cartoons and animations. In this section, we briefly describe our solutions to these challenges and the creation of a large-scale, high-quality dataset for segmentation applications.

\subsection{Preparation of Foregrounds and Backgrounds}

The initial stage of dataset synthesis can be challenging due to the difficulty of manually extracting and labeling subject instances. However, we find that chroma keying materials, as depicted in Fig.~\ref{fig:chroma_key} (a), could serve as an adequate source of such instances. These materials typically feature animated characters or objects against a keying color background, to facilitate the reuse or expansion of animations via professional software. From a chroma-keying frame (Fig. \ref{fig:chroma_key} (a)), the keying color can be automatically identified (Fig. \ref{fig:chroma_key} (b)\&(c)) and used as an alpha channel. This process allows for the extraction of background-free characters or objects (Fig. \ref{fig:chroma_key} (d)) and their pixel-based binary masks (Fig. \ref{fig:chroma_key} (e)). Please refer to the supplemental material for the detailed algorithms in data pre-processing, detection of the keying color, and the extraction of instance masks. 
In general, we collect 1,064 chroma-keying videos from popular online anime community sites, such as Niconico, Bilibili, and YouTube. To expand the foreground instances, we add still illustration subjects from the Danbooru2021 dataset \cite{danbooru2021}, in addition to those extracted from chroma-keying materials. We select illustrations with one subject instance and a clear labeling mask, denoted with \textit{solo} and \textit{transparent\_background} tags. In total, our collection comprises 65,752 subjects, where 26,844 of them from chroma-keying videos and 38,908 are from Danbooru2021's still illustrations.
For background preparation, the quality and variety of the collected background images play a crucial role for the dataset synthesis. Owing to the focus on segmentation, motion-continuous backgrounds are unnecessary. Consequently, we opt for high-quality anime stills, known as \emph{CG backgrounds}, as our image set. Our collection comprises 8,163 scenic backgrounds from the Danbooru2021 dataset, filtered by \textit{scenery} and \textit{no\_humans} tags, and 8,057 cartoon backgrounds from the bizarre pose estimator~\cite{chen2022bizarre}. The vast majority (>80\%) of images in our dataset exceed a resolution of $1024\times768$, satisfying our high-quality prerequisite.

\begin{figure*}[!t]
    \centering
    \includegraphics[width=0.99\linewidth]{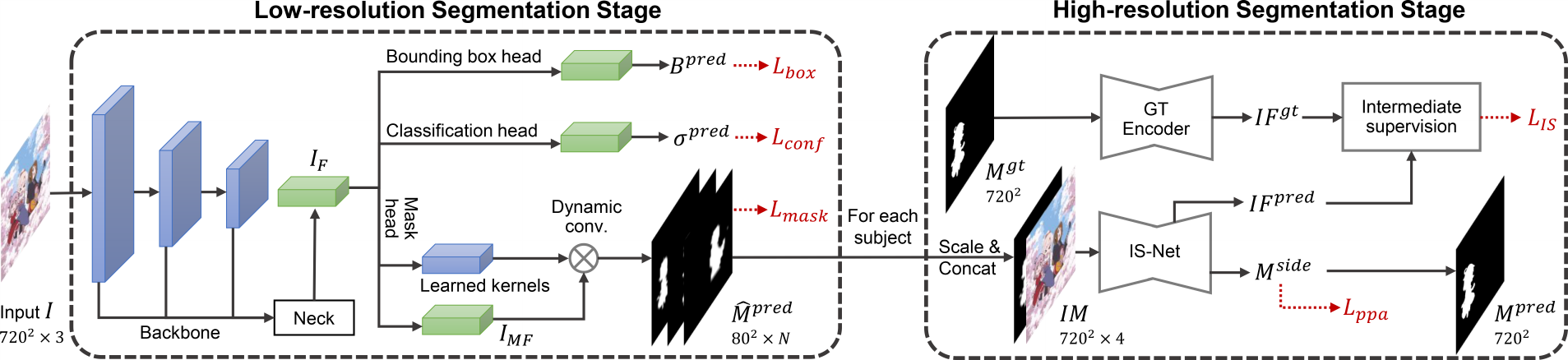}
    \caption{Model architecture overview. The low-resolution stage extracts input features and learns to predict the bounding boxes and coarse instance masks. Subsequently, high-resolution segmentation refines the coarse masks, resulting in final high-resolution subject instance extraction. Supervision objectives are highlighted in red.}
    \label{fig:model_arch}
\end{figure*}

\subsection{Dataset Synthesis}
After obtaining the appropriately isolated foreground and background elements, we proceed with the synthesis of cartoon-like images and segmentation labeling from these elements to form the final dataset. This process requires careful consideration of subject positioning to avoid overly sparse or crowded placements, as well as the development of an effective augmentation mechanism to generate a sufficient number of images for model training. Our dataset synthesis process, depicted in Figure~\ref{fig:composition}, is carried out with a standard canvas resolution of $720 \times 720$. For each canvas, we sample $N_{subj}$ subjects from a Poisson distribution with parameter $\lambda=2.5$  (Fig.~\ref{fig:composition} (a)\& (b)). The positioning of the $N_{subj}$ subject instances is guided by the ground truth bounding boxes from a randomly selected photo in the MSCOCO dataset~\cite{MSCOCO}, which includes $>N_{subj}$ instances labeled \emph{person}, in order to replicate the natural distribution of multicharacter photos (Fig.~\ref{fig:composition}(c)). 
The foregrounds are then overlayed on random cropped backgrounds (Fig.~\ref{fig:composition}(d)) for the final assembly (Fig.~\ref{fig:composition}(e)). We further refine the synthesis through optional post-placement processing, which mitigates color differences between the foreground instances and the background. This involves histogram matching for color harmonization and color quantization for more cartoon-like appearances (Fig.~\ref{fig:composition}(f)). Please refer to the supplemental material for more details in this composition process. 
Ultimately, we conduct the synthesis to form a high-quality cartoon instance segmentation dataset with 98.6k image-mask pairs. The dataset will be released upon the acceptance of the article.

\section{Cartoon Instance Extraction Network}
\label{sec:approach}
With our prepared large-scale cartoon segmentation dataset, we aim to extract high-quality instance-level extraction of subjects with a resolution of up to $720^2$.
To do so, we propose a novel two-stage segmentation approach. In the first stage of low-resolution segmentation, we locate a rough but occlusion-free mask for each subject instance. In the second stage of high-resolution segmentation, we obtain a high-resolution instance mask for each subject under the guidance of the rough instance mask acquired in the first stage. The overall design of the model architecture is illustrated in Fig.~\ref{fig:model_arch}.

\subsection{Low-resolution Segmentation Stage}
The model design of the low-resolution segmentation stage inherits from the RTMDet-Ins~\cite{lyu2022rtmdet} architecture to detect the subjects from the input cartoon image and obtain a coarse segmentation mask for each subject. The network model is mainly composed of three components: a detection backbone for feature extraction, a neck for feature fusion and refinement, and several detection heads to produce model output. Specifically, the detection backbone utilizes the enhanced CSPNeXt building blocks for a multi-level feature extraction. With the extracted multi-level feature, the neck performs a multi-level feature fusion~\cite{pafpn} to further enhance the features. Finally, we apply a bounding box head to predict the bounding box, a classification head to predict the confidence score of the bounding box, and a mask head to predict the masks of the subjects. The training of this network is versatile, which can be trained solely on the bounding box classification and regression or can be expanded to be supervised with instance masks with learnable dynamic convolutions.
This one-pass, anchor-free design is lightweight but still achieving a good computation and accuracy trade-off.

Given an input image $I$ at the resolution of $720^2\times3$, the detection backbone and neck are used to compute an enhanced multi-level feature $I_F$, which is further operated by two branches of heads. A branch predicts the bounding boxes of the subjects $B^{pred}$ and their confidence values $\sigma^{pred}$ with the bounding box and classification heads. The other branch predicts the instance masks $\hat{M}^{pred}$ with the mask head.

\paragraph{Bounding box classification and regression}
With the multi-scale feature $I_F$, $8,400$ candidate bounding boxes are created across three distinct grid layouts, which partition the original resolution of $720^2$ into grids of $80^2, 40^2$, and $20^2$. The bounding boxes are represented by four parameters $(top, left, bottom, right)$ relative to the cell center, and predicted through the bounding box head with a combination of convolution and normalization blocks in a regression manner.
For every candidate bounding box, a corresponding confidence score that signifies the likelihood that the box contains the subject is also computed using the classification head with another set of convolutional blocks following logistic regression.
The number of candidate bounding boxes typically outnumbers the ground truth number of subject instances. Thus, a dynamic soft label assignment strategy~\cite{lyu2022rtmdet} is utilized to pinpoint the top-$N$ best matches from the $8,400$ candidates bounding boxes, where $N$ is the ground truth number of subjects.

To supervise the predicted bounding boxes and confidence scores, two loss terms are designed. The GIoU loss~\cite{giou} is adopted to supervise the predicted bounding boxes, and is formulated as
\begin{equation}
    L_{box} = \frac{1}{N}\sum_{i=1}^{N}\left(y_i - \frac{C_i-|B^{pred}_{i} \cup B^{gt}_{i}|}{C_i}\right)
    \label{eqn:bbox_regression}
\end{equation}
where $i$ is the index of the bounding box, $B^{pred}_i$ is the set of pixels in the $i$-th predicted bounding box, $B^{gt}_{i}$ is the set of pixels in the ground truth of the $i$-th bounding box, $|\cdot|$ calculates the number of elements in a set, $C_i$ is the area of the smallest box that contains both $B^{pred}_i$ and $B^{gt}_i$, and $y_i=IoU(B^{pred}_{i},B^{gt}_{i})$ is the IoU score between the predicted and ground truth bounding boxes.

To supervise the confidence score, the quality focal loss~\cite{qualityfocal} is used and formulated as
\begin{equation}
    L_{conf} = -\sum^N_{i=1}|y_i-\sigma_i|^\beta\left((1-y_i)\log(1-\sigma_i)+y_i\log{(\sigma_i)}\right)
    \label{eqn:bbox_classification}
\end{equation}
where $\beta$ is a parameter to down-weigh the easy examples which achieve both good bounding box regression and confidence predictions. We set $\beta$ to 2 in our experiments.

\paragraph{Low-resolution instance segmentation}
The low-resolution instance masks are constructed after the bounding box regression. Specifically, an 8-channel feature $I_{MF}$ is extracted from $I_F$ using 4 convolutional layers. With the predicted $N$ bounding boxes obtained from the other branch, $N$ dynamic kernels are learned to filter $I_{MF}$ into $N$ low-resolution instance masks at the dimension of $80^2$, which together form the predicted low-level instances masks $\hat{M}^{pred}$ at the dimension of $80^2\times N$. A dice loss is used to supervise the predicted instance masks and formulated as
\begin{equation}
    L_{mask} = \frac{1}{N}\sum^N_{i=1}\left({1 - \frac{2\,\hat{M}_i^{gt}*\hat{M}_i^{pred}}{\hat{M}_i^{gt}*\hat{M}_i^{gt} + \hat{M}_i^{pred}*\hat{M}_i^{pred} + \epsilon}}\right)
    \label{eqn:instance_segmentation}
\end{equation}
where $\hat{M}^{gt}$ is the ground truth of $\hat{M}^{pred}$, $*$ is the element-wise multiplication operator, and $\epsilon$ is a small value for numerical stability. Both $\hat{M}^{gt}$ and $\hat{M}^{pred}$ are rescaled to the dimension of $320^2\times N$ bilinearly during loss computation.

To sum up, the following loss function is optimized for the first low-level segmentation stage:
\begin{equation}
    L_{stage1} = \lambda_{box}L_{box} + \lambda_{conf}L_{conf} + \lambda_{mask}L_{mask}
    \label{eqn:dethead_final}
\end{equation}
Here, $\lambda_{box}$, $\lambda_{conf}$, and $\lambda_{mask}$ are weighting factors and set to 2, 1, and 2 respectively in all our experiments.

\subsection{High-resolution Segmentation Stage}

For the second stage, we obtain a high-resolution mask sized $720^2$ for each instance from the predicted low-level mask acquired in the first stage, which is very important in acquiring precise masks for subjects with delicate details, for example, hair and accessories. 
To achieve this, we build our model on the IS-Net~\cite{isnet} design. Compared to alternatives such as U$^2$Net~\cite{u2net} and UNet~\cite{unet}, IS-Net takes advantage of a multi-scale ground truth encoder network to encode a ground truth instance mask $M^{gt}$ into a compact feature representation $IF^{gt}$. Such a design enables supervision for a much higher-resolution segmentation, further enhancing the model precision with a manageable memory budget. 

For each coarse instance mask $\hat{M}^{pred}_i$ (denoted as $\hat{M}$ in this subsection for conciseness) obtained in the first stage, we upscale it to $720^2$ and concatenate it to the input cartoon image $I$ to form a $720^2 \times 4$ input $IM$.
A high-resolution IS-Net segmentation network is used to predict a set of multi-scale intermediate features $IF^{pred}$ and also a set of side masks $M^{side}$ from the input $IM$. 
Similar with IS-Net, intermediate supervision is used to align predicted intermediate features $IF^{pred}$ with ground truth intermediate features $IF^{gt}$ obtained from the ground truth high-resolution mask $M^{gt}$. The side masks $M^{side}$ are obtained through a collection of last-level convolution block across all feature scale levels, with resolutions ranging from the coarsest to the finest. In our model, six levels ($D=1\cdots6$) of multi-scale computations are applied for $IF^{pred}$, $IF^{gt}$, and $M^{side}$.
The final model output $M^{pred}$ is the highest level $D=1$ of the side masks $M^{side}$. 

\paragraph{Intermediate supervision}
For intermediate supervision, we adopt the MSE loss for measuring the multi-scale feature differences between the ground truth intermediate features $IF^{gt}$ and the predicted intermediate features $IF^{pred}$ across all features scale levels, which is formulated as  
\begin{equation}
    L_{IS} = \sum^{6}_{D=1}\|IF^{pred}_D-IF^{gt}_{D}\|^2_2
    \label{eqn:intermediate_supervision}
\end{equation}
where $\|\cdot\|_2$ is the norm-$2$ operator.

\paragraph{Pixel position aware side mask supervision} While the intermediate supervision helps generate high-resolution segmentation masks, the boundaries of the masks may still be inaccurate without emphasizing the similarity between the predicted masks and the ground truth near the boundary pixels. To obtain high-resolution masks with precise boundaries, we propose to further utilize the pixel position aware loss~\cite{f3net} to supervise the multi-scale side masks $M^{side}$ as
\begin{equation}
L_{ppa} = \sum^{6}_{D=1}\lambda_D \cdot \widebar{WL_D}
\label{eqn:ppa}
\end{equation}
Here, $L_D=BCE(M^{side}_{D}, M^{gt}) + 1-IoU_p(M^{side}_{D}, M^{gt})$ is the pixel-wise similarity between the predicted side masks $M^{side}$ at the $D$-th level and the ground truth high-resolution mask $M^{gt}$ where $BCE$ is the binary cross entropy loss and $IoU_p$ calculates the $IoU$ values in an element-wise manner. $W = 1+5*|AvgPool2D(M^{gt})-M^{gt}|$ is a weighting matrix applied on the pixel-wise similarity to enhance the model's focus on boundaries. $AvgPool2D(\cdot)$ is a 2D average pooling convolution operator. $\lambda_D$ is the weighting factor of each side mask level. To emphasize the similarity of boundaries, we specially set $\lambda_6=5$ and the other five factors to $1$.

The final training loss for this stage can be written as
\begin{equation}
    L_{stage2} = L_{IS} + L_{ppa}
    \label{eqn:isloss_final}
\end{equation}

\subsection{Training Details}

For our dataset, we use 100k images for training, and the rest 7.5k images for validation. 
In the first stage, the model training is bootstrapped from the MSCOCO~\cite{MSCOCO} pre-trained RTMDet-L model, on four NVIDIA GeForce RTX 3090 graphics cards with a per-device batch size of $16$. Training spans $60$ epochs using the AdamW \cite{adamw} optimizer with a weight decay of $0.05$. 
In the second high-resolution segmentation stage, we train the model on the same testbed for $60$ epochs with an AdamW optimizer, with a learning rate of $0.001$ and weight decay of $0.005$. Refer to the supplemental material for more training details. 
In our experiments, we make the two models trained separatedly. We have attempted a joint training strategy but have observed little improvement. Under the joint training configuration, we found the second stage gradients can hardly enhance the detection heads of the first stage and the overall model performance as well. Furthermore, the joint training approach substantially increased resource usage, making it unsuitable for training on large-scale datasets.

\begin{table}[!t]
\begin{tabular}{l|ccc}
\hline
Model                                   & \makecell{Box\\AP~$\uparrow$} & \makecell{Mask\\AP~$\uparrow$} & \makecell{Boundary\\AP~$\uparrow$} \\ \hline
Grounded SAM (Original)                       &      92.2       &       46.7       &      16.6           \\ \hline
Grounded SAM (Finetuned)              &      92.2       &       90.1       &      49.7       \\ \hline
\makecell[l]{SOLOv2 (YAAS)}                        &      64.3       &       59.0       &      26.2          \\ \hline
SOLOv2 (Finetuned)   &      76.8      &       70.8       &      22.8                                    \\ \hline
SAM (Instance token tuning)                       &      18.7    &       6.5      &      0.2        \\ \hline
SAM (Full finetuning)              &      24.2       &       8.4       &      0.2       \\ \hline
Ours                                &      \textbf{93.1}       &       \textbf{93.2}       &   \textbf{63.6}    \\ \hline
\end{tabular}
\caption{Quantitative comparisons with existing methods.}
\label{tab:quanlitative_evaluations}
\end{table}

\section{Evaluations and Discussions}
\label{sec:evaluation}
\begin{figure*}[!t]
    \centering
    \includegraphics[width=\linewidth]{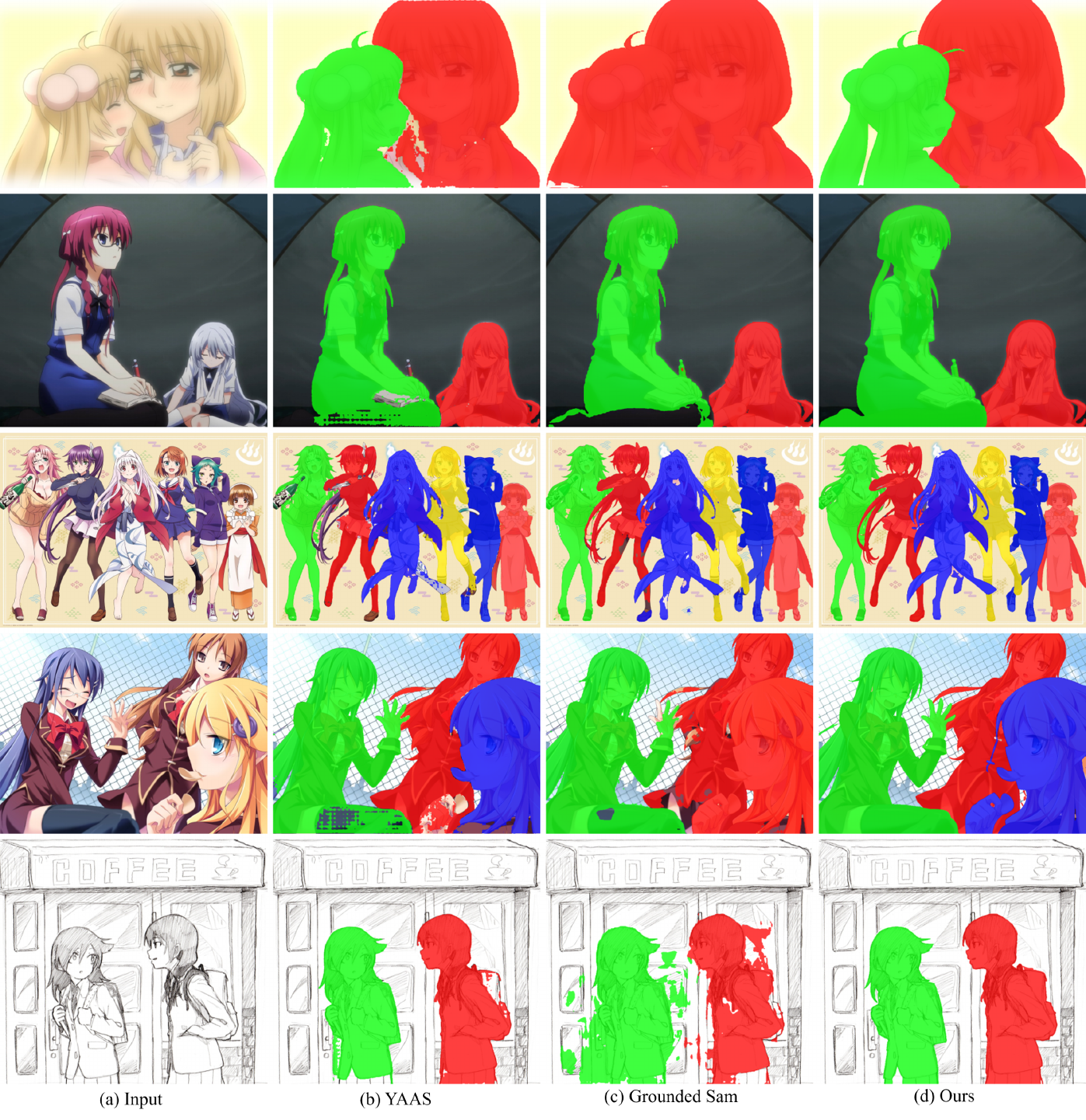}
    \caption{Qualitative comparisons with existing methods.}
    \label{fig:qualitative_evaluation}
\end{figure*}

In this section, we shall evaluate the effectiveness of our two novel components: the large-scale cartoon segmentation dataset, and the two-stage cartoon instance segmentation model. The evaluation is carried out by comparing our solutions with prevalent existing solutions, including the classic learning-based Mask R-CNN approach~\cite{maskrcnn}, the state-of-the-art photo segmentation method Grounded Segment Anything~\cite{groundedsegmentanything}, and the state-of-the-art cartoon segmentation method YAAS~\cite{yaas}. Each contender model is prepared in two variations: the original pre-trained model and a fine-tuned model using our proposed dataset, for a thorough and unbiased analysis. 
To evaluate the efficacy of our proposed large-scale dataset, we compare with these existing solutions between their pretrained versions and fine-tuned versions with our dataset, and our model undergoes an additional abalation study. On the other hand, when evaluating the model design, we only take the finetuned models into account during comparison. 
We conduct both quantitative and qualitative evaluations, with all analysis conducted on a separatedly prepared test dataset. This dataset comprises 307 cartoon and manga samples across a wide array of cartoon, illustration, and manga, with diversity in visual content and style, all of which have been manually labeled for ground truth segmentation. We have ensured that none of the models has previously seen the samples contained within this test dataset during training.


\begin{figure*}[!t]
    \centering
    \includegraphics[width=\linewidth]{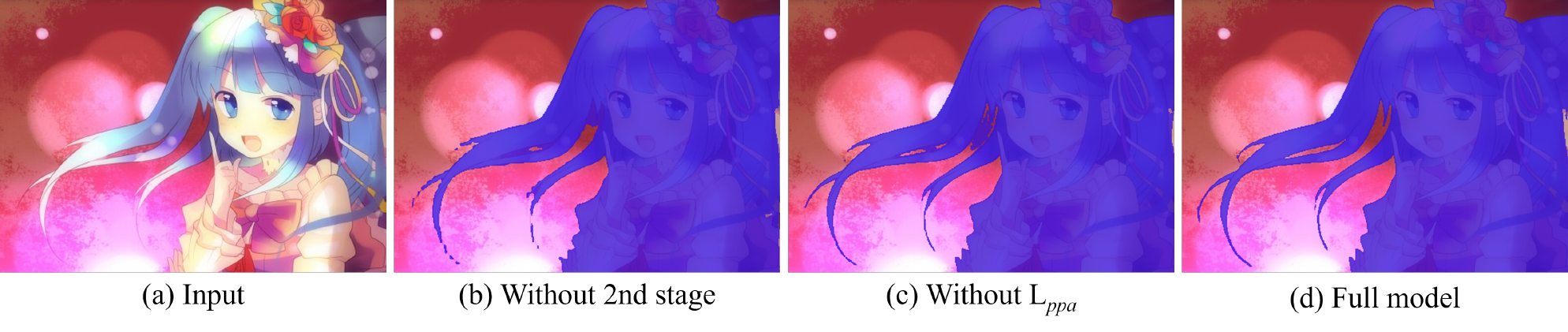}
    \caption{Ablation studies. (a) Input. (b) The absence of the second high-resolution segmentation stage results in inaccurate masks. (c) Without loss of $L_{ppa}$ in high-resolution segmentation training, mask boundary accuracy is compromised. (d) Our full model exhibiting superior quality.}
    \label{fig:ablation}
\end{figure*}

\begin{table}[!t]
\begin{center}
\begin{tabular}{ l | c c c}
\hline
Method & Box AP~$\uparrow$ & Mask AP~$\uparrow$ & Boundary AP~$\uparrow$ \\ \hline
\makecell[c]{1/8 dataset} &  89.4 & 83.3 & 44.2 \\
\makecell[c]{W/o 2nd stage}     & \textbf{93.1} & 84.6 & 52.1 \\
W/o $L_{ppa}$         & \textbf{93.1} & 92.9 & 63.1 \\
Full model & \textbf{93.1} & \textbf{93.2} & \textbf{63.6}  \\
\hline
\end{tabular}
\caption{Quantitative ablation studies on the proposed model and dataset.}
\label{tab:ablation}
\end{center}
\end{table}

\subsection{Quantitative Evaluations}


We first conduct a quantitative evaluation to compare our method and all competitors with three evaluation metrics: Box AP, Mask AP, and Boundary AP~\cite{boundaryiou}. 
Among these metrics, Box AP computes the precision of the predicted bounding boxes against their corresponding ground truths; Mask AP measures the precision of the instance masks, and Boundary AP computes the precision over the instance boundary areas. Given the ground truth segmentation mask $G$ and the predicted mask $P$, their boundary IoU is formulated as:
\begin{equation}
  \text{Boundary IoU}(G, P) = \frac{|(G_d \cap G) \cap (P_d \cap P)|}{|(G_d \cap G) \cup (P_d \cap P)|}
  \label{eqn:eqn_boundary}
\end{equation}
where boundary regions $G_d$ and $P_d$ are the sets of all pixels within the distance of $2\%$ of the image diagonal length from their exterior mask contours. We reveal the evaluation statistics in Table~\ref{tab:quanlitative_evaluations}.

We first assess the effectiveness of our proposed large-scale dataset. Upon examination, we observe that fine-tuning on this large-scale dataset significantly enhances the performance across all existing models. We hypothesize that the distinctive features of cartoons and animations in our dataset enhance the learning-based solutions' ability to handle the special delicate styles more proficiently. Furthermore, we conducted an ablation study on our proposed model. We compared the full model with an alternative which was exposed to only $1/8$ of the total dataset size. As indicated in Table~\ref{tab:ablation}, we believe a substantial cartoon-specific dataset is indispensable and cannot be easily replaced by a few-shot alternative. 



We afterward examine the systematic design of different models. Among the three competitors, Mask R-CNN presents the lowest performance attributable to the significant number of unidentified subjects, reflected in its low Box AP scores. The model is significantly limited by its architecture design where high-resolution features are not fully utilized during inference. On the other side, the transformer-based model grounded SAM achieves better performance in both subject identification and segmentation with our large-scale dataset, but still cannot reach on-par performance as our proposed model. Additionally, the model features a lower native segmentation resolution compared to ours ($256^2$ vs. $720^2$), which leads to a significantly weaker ability to extract instance boundaries. Furthermore, their method is not fully automatic and requires a manual stage of bounding box extraction from the text prompt. In contrast, our proposed model is designed to work fully automatic on cartoon subject instances. The SOLOv2 model, whether trained with the YAAS dataset~\cite{yaas}, or with our large-scale dataset, still cannot surpass our performance when trained on our model. We believe this is due to their relatively smaller model capacity and the lack of flexible image feature extraction modules such as CSPNeXt and the learning of feature fusions. 
We also explore the possibilities of directly expanding the original SAM model to recognize cartoon subject instances. This could be done by fine-tuning the SAM mask decoder model with  additional instance tokens or the complete SAM pipeline, which includes the finetuning of the image encoder as well. During inference, we shall directly use the learned instance tokens as prompts for segmentation. However, according to statistics, both solutions perform poorly, which we believe is due to the vast model space of the SAM model, which complicates transfer learning and fine-tuning. We observe that the model is susceptible to overfitting, even when trained on our large-scale dataset comprising more than 100k images.

\begin{figure*}[!t]
    \centering
    \includegraphics[width=\linewidth]{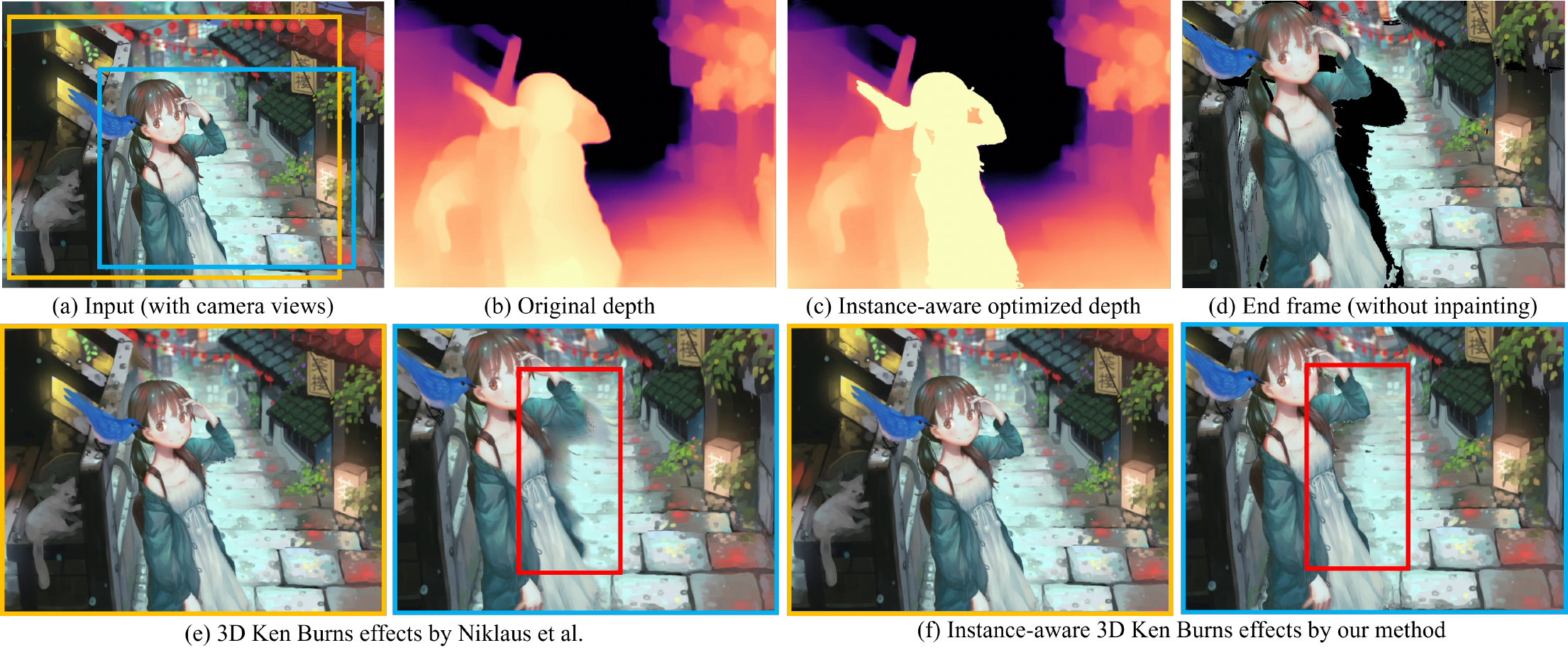}
    \caption{3D Ken Burns synthesis with instance awareness. Our extraction of instances aids in optimized geometry estimation and visually pleasing inpainting for seamless view interpolation and synthesis.}
    \label{fig:3dkenburns}
\end{figure*}

\begin{figure*}[!t]
    \centering
    \includegraphics[width=\linewidth]{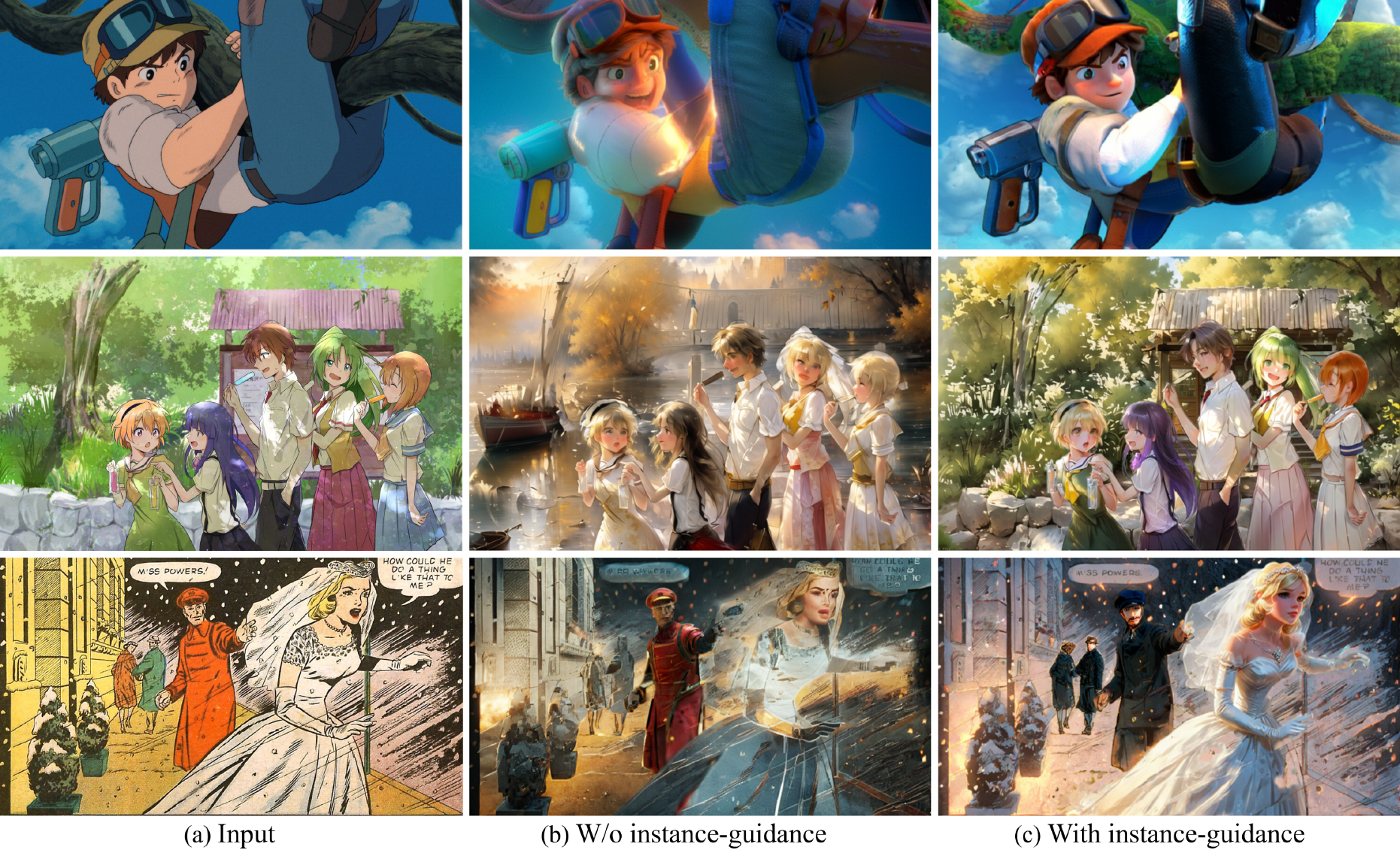}
    \caption{Instance-aware style editing with global prompt \textit{3D pixar style} (top), \textit{artist\_cg, cg} (middle), \textit{cinematic, ultra detailed} (bottom). Our style editing solution respects the target style prompt and maintains the visual semantics for all instances. }
    \label{fig:style_editing}
\end{figure*}

\subsection{Qualitative Evaluations}

We present our qualitative visual comparison in Fig.~\ref{fig:qualitative_evaluation}. In this evaluation we do not compare with Mask R-CNN due to its large margin to the state-of-the art in performance. 
According to Fig.~\ref{fig:qualitative_evaluation} (b), the SOLOv2-based segmentation model YAAS is relatively weak in extracting image features, causing incomplete extraction of instances in almost all sample cases. Moreover, the extraction may suffer from checkerboard visual artifacts. According to our experiments, our dataset helps the SOLOv2 model to better deal with the parsing of croweded images with multiple instances, yet the overall performance is still suboptimal. 
On the other hand, as in Fig.~\ref{fig:qualitative_evaluation} (c), the fine-tuned Grounded SAM model manages to extract all instances in our demonstrated samples, but may fail to distinguish individual characters if they do not share clear boundaries (top row) and may cause tear-like visual artifacts (2nd, 5th rows). Moreover, the SAM model cannot extract precise instance boundaries, especially when the instance features delicate drawings such as long hair, fancy clothes, or accessories. Additionally, the model does not generalize well on abstract line drawings and manga images. 
In contrast, our method, supported by a large-scale cartoon segmentation dataset and a high-resolution segmentation model tailored for cartoon images, successfully identifies all subjects with high-quality masks in all images tested. Particularly, we emphasize our method's effectiveness in challenging scenarios, such as large areas of occlusion, intricate structures at the detail level, and adaptability to various styles of cartoon-style images.

\subsection{Ablation Studies}
To verify the value of the large-scale dataset for our proposed model design, we first conduct an ablation study comparing the training results using the full training dataset and a subset equivalent to 1/8 of the full data. Table~\ref{tab:ablation} shows that a constrained dataset reduces the overall performance of the model, again validating the importance of creating this large-scale dataset. We further validate our model's design by respectively removing the second stage and the pixel position aware loss $L_{ppa}$. Fig.~\ref{fig:ablation} presents the visual results of this ablation study. Observably, without the second high-resolution segmentation stage, the model fails to produce accurate masks for intricate structures such as hairs. Incorporating the high-resolution segmentation stage substantially improves the quality of the segmented instance masks. Without the pixel position aware loss $L_{ppa}$, the model still struggles to capture precise structures and mask boundaries. The qualitative statistics also support these specific model designs.

\section{Applications}
\label{sec:application}
The high-quality instance-level extraction of cartoon subjects enables various cartoon editing applications. These were previously considered challenging and time-consuming due to the significant reliance on manual segmentation of scenes and characters. In this section, we introduce three major applications essential to the real-world production of cartoons. 

\subsection{3D Ken Burns}
The 3D Ken Burns \cite{3dkenburns}, or in general the parallax effect, has found wide use in low-cost cartoon animation productions due to its ability to create dynamic visuals from a few numbers of static frames, eliminating the need for complex animation production pipeline. 
The original version of 3D Ken Burns uses a semantic-aware neural network to estimate the depths of the scene. The depth information of the input image guides the rendering of the scene from other camera's positions and angles. Disocclusions are addressed with context-aware color and depth inpainting to achieve coherent synthesis results. 
However, the creation of a satisfactory 3D Ken Burns requires comprehensive and detailed understanding of the geometry and texture of the scenes to guide the rendering of frames for camera movement. The current model, as in Fig.~\ref{fig:3dkenburns} (e), struggles to estimate the depth and textures around the subject instances due to the lack of awareness of the subject instance, causing obvious visual artifacts. later works such as~\cite{comicdepth} attempts to refine the depths of cartoon scenes, but they are still not ready for a visually consistent rendering.

To cope with ambiguity in depth estimation, we propose to first estimate an initial depth with a refined depth estimator~\cite{yin2022towards}. After that, we apply an average pooling filter on the initial depth of each subject instance. As in Fig.~\ref{fig:3dkenburns} (c), our depth estimation better differentiates between the foreground and the background. 
On the other hand, we also leverage instance awareness to improve efficiency and reduce ambiguity for the inpainting of the occluded areas during scene rendering. This is achieved by isolating the foreground information from the complete scene, enabling us to retrieve the missing texture information exclusively from the background. In our implementation, we utilize latent diffusion~\cite{latentdiffusion} to perform high-quality and consistent inpainting on the foreground instances and the occluded pixels (Fig.~\ref{fig:3dkenburns} (d)). For each rendered frames to be inpainted, we first fill the inpainting area using PatchMatch~\cite{patchmatch} with patch size $=3$ as an initialization. After that, we input this roughly inpainted image with the SwinV2 Tagger~\cite{swinv2tagger} model to compute a set of visual tags for the background, which we use as an additional constraint to ensure consistency in visual semantics. Besides, we also design a fixed negative prompt, \textit{human, single, person, girl, 1girl, boy, 1boy, creature, animal, alien, robot, body}, to restrict the synthesis of these specific subject visual semantics. We then perform noise application with a strength (e.g., the style translation strength) of 0.75, and denoise it with the DPM++ 2M Karras scheduler with 32 diffusion steps.
As in Fig.~\ref{fig:3dkenburns} (e), we can achieve more engaging and immersive visual experiences of 3D Ken Burns, thanks to instance-aware optimization steps for both the geometry and texture during the rendering process.

\begin{figure*}[!t]
    \centering
    \includegraphics[width=\linewidth]{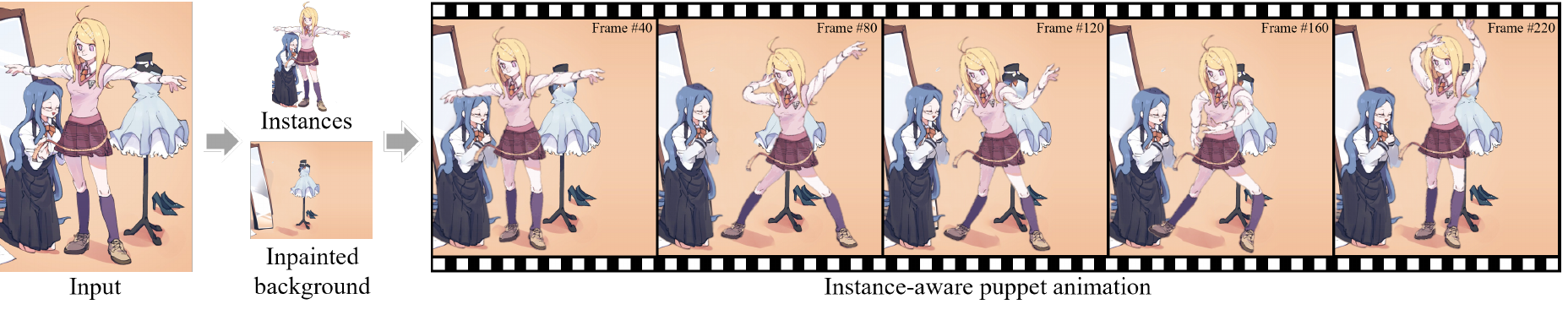}
    \caption{We can synthesize puppet animations from the segmented cartoon instances. Without our proposed pipeline of instance segmentation and background inpainting, the manual production of these animations will be labor-intensive. }
    \label{fig:motion_manga}
\end{figure*}

\begin{figure}[!t]
    \centering
    \includegraphics[width=\linewidth]{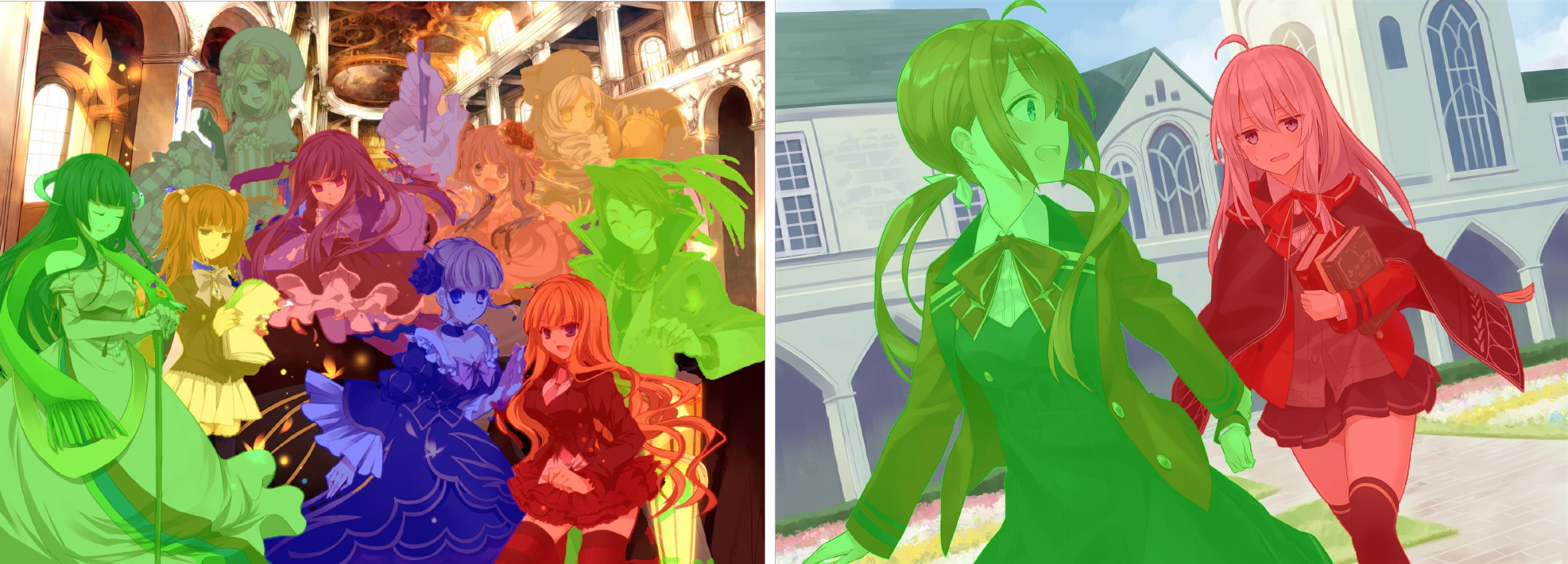}
    \caption{Limitations of our approach. The segmentation may be confused when the image content is too crowded, or contains major occlusion.}
    \label{fig:limitations}
\end{figure}

\subsection{Instance-aware style editing}

The technique of image-to-image style editing and translation is rapidly progressing, and some existing solutions have shown potential in translation between different styles, such as between Japanese anime and western comics~\cite{screenvae} or from photo to anime styles~\cite{cartoongan}, they cannot accomplish the objective of style translation across arbitrarily specified visual styles. The solutions to this challenging task is expected to have both a high-level semantic understanding of the visual contents, as well as the ability to synthesize high-quality textures with respect to the semantics. 
Recent trends in style editing favor diffusion models \cite{ddim, ddpm} due to their superior generation quality, flexibility in image generation, and integration with zero-shot generation capabilities of language models \cite{latentdiffusion}. In practical terms, given an input image and a description of the target style, these are encoded using the CLIP text encoder~\cite{radford2021learning} and used to condition the denoising process for the synthesis of the target style through a cross-attention module. However, in the absence of instance-level guidance during image style editing, these models often fail to maintain the identity and visual semantics of the instances. As illustrated in Fig.~\ref{fig:style_editing} (b), we note that instances may lose their original visual properties, such as hair colors and background semantics (2nd row), face expressions (1st row), and may not fully adhere to the style prompts (3rd row) when the style translation is guided solely by a global style prompt. We hypothesize that this visual discrepancy problem originates from the indiscriminate nature of CLIP embedding injection during the denoise diffusion step, which causes confusion in applying proper prompts to only a selection of visual elements. Additionally, a single global style prompt may not provide sufficient guidance for classifier-free guidance, resulting in a minor decline in visual quality, such as blurry edges (1st row) and halo artifacts (2nd row).  

We propose tackling these visual discrepancy issues in cartoons with a consistent style translation using a fine-grained instance-level diffusion method. This method is two-fold: firstly, we extract each subject instance and calculate its visual tags with the SwinV2 tagger~\cite{swinv2tagger} to capture the instance-level semantics. We then implement the instance-level image-to-image diffusion-based translation for each instance, combining the global style prompt with instance-specific visual tags to create the final style translation prompt. It is important to note that we also calculate the inpainted background from the subjects and treat it as a unique instance to maintain visual similarities after translation. Furthermore, we utilize the lineart ControlNet~\cite{zhang2023adding} to ensure that the synthesized image maintains the same structural lines as the original, offering better control over local details. In our experiments, all instances receive the same noise seed and schedule for style consistency. We adjust the style editing strength to 0.75, with a total of 24 denoising steps under the DPM++ 2M Karras denoise scheduler. The ControlNet guidance strength is set to 1. We demonstrate our results in Fig.~\ref{fig:style_editing} (c). From the results, we can observe that our results enjoy a clearer and sharper visual impression. Crucially, our style editing maintains the visual semantics of the subject instances while adhering to the global style target prompts. Notably, the advent of temporal consistent video editing~\cite{ebsynth, guo2023animatediff} allows our solution to extend to videos without the need for manual editing of each frame. We provide further results of this style editing application in the supplemental material.

In addition to visual demonstrations, we also performed a quantitative assessment using the CLIP score~\cite{radford2021learning} to measure the style translation results against the global text prompt. A greater CLIP score implies a higher conformity of the style translation to the text description of the target styles. In this experiment, our proposed solution achieves a CLIP score of 26.8, as opposed to the 23.9 score obtained by the style translation using a singular global style prompt. This further proves that our instance-aware style editing approach reduces the confusion lying inside the CLIP model during target style injection. This further substantiates that our instance-aware style editing approach minimizes confusion during target style injection and consequently results in a more consistent translation.

\subsection{Puppet Animation from Illustration/Manga}

The extraction of high-resolution subject instances also allows the creation of puppet animation from a single illustration or manga image. Without individual masks for each instance, one can only apply point-based or grid-based deformation to the overall image, which usually leads to badly deformed subjects. Additionally, semantics-based transformation is not feasible without instance masks, such as pose-based transformation. 
We propose a conceptual approach for generating puppet animation, as depicted in Fig.~\ref{fig:motion_manga}. The process begins with the extraction of individual instances, followed by the application of techniques presented in Animated Drawings~\cite{animated_drawings} to create high-quality, time-varying deformation frames of the subject. To manage any missing pixels due to warping, we use the same inpainting technique employed in the 3D Ken Burns application, ensuring that no additional subjects are created during inpainting. Our proposed method allows for the creation of visually coherent puppet-animation, either from a reference animation or through manual manipulation of image keypoints. Crucially, the movement of each subject operates independently, without any impact on other subjects or the background.


\section{Discussion and Conclusion}
\label{sec:conclusion}


While our method offers promising results, there are certain aspects that need further refinement. One potential issue arises when handling images with a high degree of subject crowding, particularly if the subjects share similar colors and styles. As shown in the left image of Fig.~\ref{fig:limitations}, our method can struggle to identify the belonging of the dark pixels in the lower middle part. Furthermore, occlusions can lead to discontinuity in the subject area, which may result in small, detached areas being misidentified as part of the occluding subject. For example, in the right image of Fig.~\ref{fig:limitations}, the right hand of the girl in the back is incorrectly identified as part of the girl in the front. We believe that including training examples with occlusions could mitigate such issues, but generating such examples poses a challenge.  
It is also worth noting that our model primarily identifies human characters as subject instances because of the way our dataset is prepared. Despite this bias, the model is capable of extracting other non-human instances, e.g., cats, pigeons, and robots. In case fully automatic extraction of these instances may not always be successful, one can still achieve semi-automatic extractions by providing an additional instance bounding box, as the second segmentation stage can be initiated with any bounding box prompts. Considering that an ultimate SAM-based solution is technically infeasible for cartoons, we maintain that our model strikes a balance between generalization, output quality, and cost.

\paragraph{Conclusion} For facilitate the cartoon editing pipelines that require segmentation, we proposed a new large-scale cartoon segmentation dataset and a novel cartoon-tailored segmentation model that can produce high-quality instance masks for the subjects in a cartoon image. Especially, our dataset is prepared using a novel reverse engineering approach based on chroma-keying videos, still illustrations, and manga images. Our method shows significant superiority over existing methods in both qualitative and quantitative evaluations, and leads to high-quality cartoon editing applications, including 3D Ken Burns effect, cartoon style editing, and puppet animations.

\bibliographystyle{abbrv-doi-hyperref}

\bibliography{sample-base}

\end{document}